# Ergodic Annealing[*]

Carlo Baldassi, Fabio Maccheroni, Massimo Marinacci, Marco Pirazzini

Bocconi University

August 1, 2020

**Abstract**

Simulated Annealing is the crowning glory of Markov Chain Monte Carlo Methods for the solution of NP-hard optimization problems in which the cost function is known. Here, by replacing the Metropolis engine of Simulated Annealing with a reinforcement learning variation – that we call Macau Algorithm – we show that the Simulated Annealing heuristic can be very effective also when the cost function is unknown and has to be learned by an artificial agent.

# 1 Introduction

The recent years and events lead to a massive development of content-oriented cloud services. The most popular and voluminous content offered in today's networks are videos that must be efficiently delivered to end customers. The objective of the service provider (root) is to optimize the delivery of content to its costumers (terminals). In this optimization problem the cost is usually assumed to be known (left graph). Yet, in reality it is often unknown because it depends on many stochastic factors, such as the traffic on the network, the level of demand, and so on (right graph).

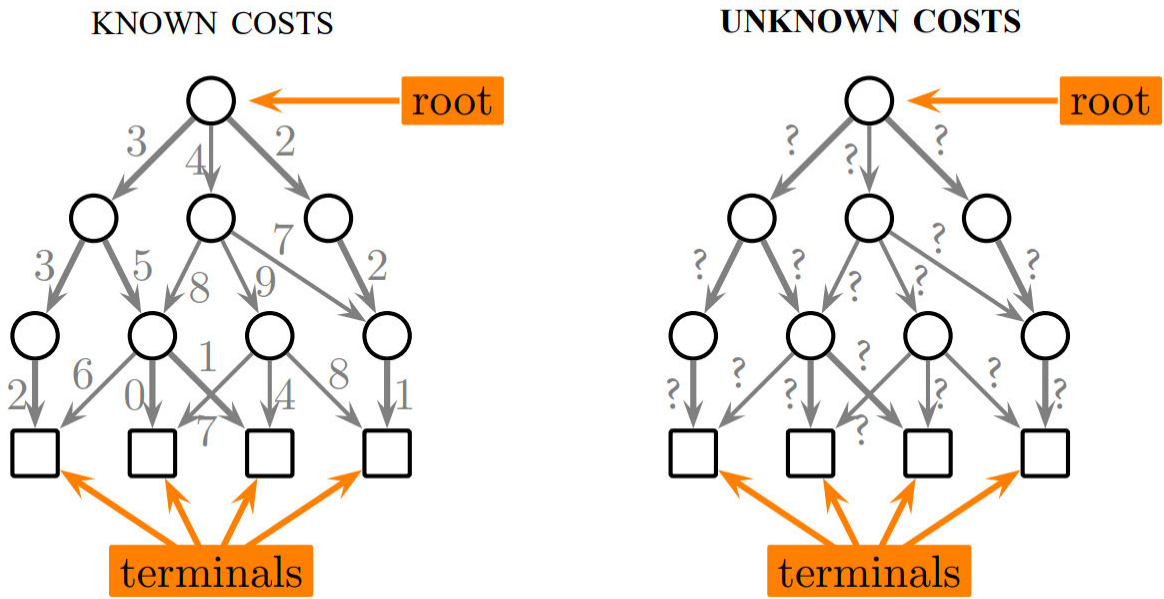

Figure 1: Graphical representation of networks where information travels from a root to a set of terminals over channels with known or unknown cost.

[*]We thank Simone Cerreia-Vioglio for very useful discussions as well as the ERC (grant INDIMACRO) and PRIN (grant 2017CY2NCA) for financial support.

This is just an instance of a general decision problem in which, *ex ante*, the Decision Maker (DM) ignores the payoff of the available actions and has limited resources to discover it. In this note, we show that a natural modification of the Simulated Annealing algorithm of Kirkpatrick et al. (1983) permits to efficiently solve this conceptually non-trivial problem.

## 2 Known payoffs: Metropolis Algorithm and Simulated Annealing

Let $A$ be a finite set of actions, $u : A \to \mathbb{R}$ be a **known** objective function that the DM aims to maximize. When the number of alternatives is small, the DM can just use a brute force comparison-and-elimination algorithm that, after $|A| - 1$ binary comparisons, finds the optimal alternative.

When the number of alternatives increases, one needs to go beyond this basic algorithm. In particular, one can rely upon the celebrated *Metropolis Algorithm* (Metropolis et al., 1953) and its evolution called *Simulated Annealing* (Kirkpatrick et al., 1983).

---

**Metropolis Algorithm** Let $\beta > 0$.

**Step 0.** Choose $a_0 \in A$ randomly.[1]

**Step $n + 1$.** Choose $b \in A$ randomly.[1]

- If $u(b) \geq u(a_n)$, then accept $b$ as $a_{n+1}$.

- If $u(b) < u(a)$, then

  ○ accept $b$ as $a_{n+1}$ with probability $e^{\beta[u(b)-u(a_n)]}$;

  ○ reject $b$ and maintain $a_n$ as $a_{n+1}$ with probability $1 - e^{\beta[u(b)-u(a_n)]}$.

---

This is the Metropolis Algorithm with *inverse temperature* $\beta$, *initial distribution* $\mu$ and *proposal matrix* $Q$.[1] Its key property is that, if the current state is $a$, the next state $b$ is determined according to the following transition probabilities

$$P(b \mid a) = \begin{cases} Q(b \mid a) \min\left\{1, e^{\beta[u(b)-u(a)]}\right\} & \text{if } b \neq a \\ 1 - \sum_{c \in A \setminus \{a\}} Q(c \mid a) \min\left\{1, e^{\beta[u(c)-u(a)]}\right\} & \text{if } b = a \end{cases}$$

Thus, the algorithm realizes an aperiodic and irreducible Markov chain with stationary distribution

$$p_\beta(a) = \frac{e^{\beta u(a)}}{\sum_{b \in A} e^{\beta u(b)}} \quad \forall a \in A$$

---

[1] Here, $\mu \in \Delta(A)$ is the initial distribution, which is used to randomly select $a_0$ in Step 0 of the algorithm; while $Q(\cdot \mid a_n) \in \Delta(A)$ is the Markovian distribution, which is used to randomly select $b$ in Step $n + 1$ of the algorithm, and depends on the incumbent $a_n$.

Specifically, $Q(\cdot \mid a_n)$ corresponds to the $a_n$-th row of a symmetric and irreducible $A \times A$ stochastic matrix $Q$. This matrix describes the way in which the algorithm explores the landscape $A$. Irreducibility guarantees full exploration of $A$, symmetry is intuitive and it can be dispensed with. See Hastings (1970) or Madras (2002) for a textbook treatment.

Therefore, when the space $A^{\mathbb{N}}$ of all infinite sequences $\boldsymbol{a} = \{a_0, a_1, ...\}$ of elements of $A$ is considered, the Ergodic Theorem guarantees that

$$\mathbb{P}\left[\boldsymbol{a} \in A^{\mathbb{N}} : \lim_{n \to \infty} \frac{\delta_a(a_0) + \delta_a(a_1) + ... + \delta_a(a_n)}{n+1} = \frac{e^{\beta u(a)}}{\sum_{b \in A} e^{\beta u(b)}}\right] = 1 \qquad \forall a \in A$$

That is, the long-run frequency with which $a$ is chosen from $A$ is almost surely $p_\beta(a)$.[2]

The idea of Simulated Annealing is to slowly increase $\beta$, while the Metropolis algorithm runs, with the objective of approaching the limit distribution

$$p_\infty(a) = \frac{1}{|\arg\max_A u|} \delta_a(\arg\max_A u) \qquad \forall a \in A$$

In the words of its creators: "At each temperature [here $1/\beta$], the simulation must proceed long enough for the system to reach a steady state." Thus, $\beta_0$ should be maintained for $n = 0, ..., t_0 - 1$, with $t_0$ large enough to achieve the stable (empirical) frequency

$$\hat{p}_0(a \mid \boldsymbol{a}) = \frac{\delta_a(a_0) + \delta_a(a_1) + ... + \delta_a(a_{t_0})}{t_0 + 1} \approx \frac{e^{\beta_0 u(a)}}{\sum_{b \in A} e^{\beta_0 u(b)}} \qquad \forall a \in A$$

Subsequently, $\beta_1 > \beta_0$ should be maintained for $n = t_0, ..., t_1 - 1$, with $t_1$ large enough to achieve the stable (empirical) frequency

$$\hat{p}_1(a \mid \boldsymbol{a}) = \frac{\delta_a(a_0) + \delta_a(a_1) + ... + \delta_a(a_{t_1})}{t_1 + 1} \approx \frac{e^{\beta_1 u(a)}}{\sum_{b \in A} e^{\beta_1 u(b)}} \qquad \forall a \in A$$

and so on, aiming at a long-run frequency

$$\lim_{k \to \infty} \hat{p}_k(a \mid \boldsymbol{a}) \approx \lim_{k \to \infty} \frac{e^{\beta_k u(a)}}{\sum_{b \in A} e^{\beta_k u(b)}} = \frac{1}{|\arg\max_A u|} \delta_a(\arg\max_A u) \qquad \forall a \in A$$

The sequence $\{t_k, \beta_k\}_{k \in \mathbb{N}}$ is called *annealing schedule*.[3]

---

**Simulated Annealing** Let $\{t_k, \beta_k\}_{k \in \mathbb{N}}$ be an annealing schedule.

**Step 0.** While $n < t_0$ perform the **Metropolis Algorithm** with inverse temperature $\beta_0$.[4]

**Step $k+1$.** While $t_k \leq n < t_{k+1}$ perform Step $n+1$ of the **Metropolis Algorithm** with inverse temperature $\beta_{k+1}$.[5]

---

The algorithm runs until the system "freezes", that is, $a_n$ stops changing (or until a given number $N$ of iterations has been preformed). The selected alternative is a candidate maximizer for the objective function $u$.[6]

---

[2] Here $\mathbb{P}$ is the Markovian probability $\mathbb{P}[\{a_0\} \times \{a_1\} \times ... \times \{a_n\} \times A \times A \times ...] = \mu(a_0) \prod_{i=1}^{n} P(a_i \mid a_{i-1})$ for all $n \in \mathbb{N}$ and all $(a_0, a_1, ..., a_n) \in A^{n+1}$. Note that $p_\beta$ is also the asymptotic distribution of coordinates of the Markov chain, that is, $\mathbb{P}\left[\boldsymbol{a} \in A^{\mathbb{N}} : a_n = a\right]$ converges to $p_\beta(a)$ as $n \to \infty$, for all $a \in A$.

[3] Formally, $\{t_k\}_{k \in \mathbb{N}}$ is a strictly increasing sequence of strictly positive integers and $\{\beta_k\}_{k \in \mathbb{N}}$ is a strictly increasing diverging sequence in $(0, \infty)$. Originally, $t_k = (k+1)L$ for some fixed "large" loop lenght $L \in \mathbb{N}^*$ and $\beta_k = (1+\rho)^k \beta_0$ for some "small" factor $\rho \in (0, \infty)$.

[6] This generates $a_0, a_1, ..., a_{t_0}$.

[7] This generates $a_{t_k+1}, a_{t_k+2}, ..., a_{t_{k+1}}$.

[6] See Gelfand and Mitter (1985, 1987) and the review of Romeo and Sangiovanni-Vincentelli (1991) for conditions on the annealing schedule that guarantee convergence of Simulated Annealing to $p_\infty$.

# 3 Unknown payoffs: Macau Algorithm and Ergodic Annealing

Now assume that the objective function $u$ is **unknown** to the DM. In particular, we consider the important case when $u(a)$ is the expectation of the random payoff $U(a)$ of alternative $a \in A$. This objective function may be unknown because, for example, the DM ignores the distribution $F_{U(a)}$ of the random variable $U(a)$.

Denoting by $\{U_n(a)\}_{n \in \mathbb{N}^*}$ a process consisting of i.i.d. copies of $U(a)$, the payoff of the DM if she chooses action $a$ at period $n$ is the (observable) realization

$$v_n(a)$$

of the random variable $U_n(a)$. A function $u_0 : A \to \mathbb{R}$ represents the *ex ante* evaluation of the DM, and we set $U_0(a) \equiv u_0(a)$.

---

**Macau Algorithm** Let $\beta > 0$.

**Step 0.** Choose $a_0 \in A$ randomly.

**Step $n+1$.** Choose $b \in A$ randomly.

- If $u_n(b) \geq u_n(a_n)$, then accept $b$ as $a_{n+1}$.
- If $u_n(b) < u_n(a_n)$, then

   ○ accept $b$ as $a_{n+1}$ with probability $e^{\beta[u_n(b) - u_n(a_n)]}$;

   ○ reject $b$ and maintain $a_n$ as $a_{n+1}$ with probability $1 - e^{\beta(t)[u_n(b) - u_n(a_n)]}$.

Observe $v_{n+1}(a_{n+1})$ and set

$$C = \delta_{a_{n+1}}(a_0) + \delta_{a_{n+1}}(a_1) + \cdots + \delta_{a_{n+1}}(a_{n+1}) \quad [\text{number of times } a_{n+1} \text{ has been chosen}]$$

$$u_{n+1}(a_{n+1}) = \frac{C-1}{C} u_n(a_{n+1}) + \frac{1}{C} v_{n+1}(a_{n+1}) \quad [\text{updated empirical average of } U(a_{n+1})]$$

$$= \frac{\sum_{t \in \{0,\ldots,n+1\}: a_t = a_{n+1}} v_t(a_t)}{\delta_{a_{n+1}}(a_0) + \delta_{a_{n+1}}(a_1) + \cdots + \delta_{a_{n+1}}(a_{n+1})}$$

and $u_{n+1}(a) = u_n(a)$ for all $a \in A \setminus \{a_{n+1}\}$.

---

Heuristically, this algorithm should lead to the softmaximization of $u$ with inverse temperature $\beta$.

ERGODIC CONJECTURE

$$\mathbb{P}\left[\boldsymbol{a} \in A^{\mathbb{N}} : \lim_{n \to \infty} \frac{\delta_a(a_0) + \delta_a(a_1) + \ldots + \delta_a(a_n)}{n+1} = \frac{e^{\beta u(a)}}{\sum_{b \in A} e^{\beta u(b)}}\right] = 1 \qquad \forall a \in A$$

ASYMPTOTIC CONJECTURE

$$\lim_{n \to \infty} \mathbb{P}\left[\boldsymbol{a} \in A^{\mathbb{N}} : a_n = a\right] = \frac{e^{\beta u(a)}}{\sum_{b \in A} e^{\beta u(b)}} \qquad \forall a \in A$$

We are now ready to perform Simulated Annealing using Macau rather than Metropolis to find the optimal action when the DM ignores the objective function, that is, the true expected payoff $u$.

---

**Ergodic Annealing** Let $\{t_k, \beta_k\}_{k \in \mathbb{N}}$ be an annealing schedule.

**Step 0.** While $n < t_0$ perform the **Macau Algorithm** with inverse temperature $\beta_0$.

**Step $k+1$.** While $t_k \leq n < t_{k+1}$ perform Step $n+1$ of the **Macau Algorithm** with inverse temperature $\beta_{k+1}$.

---

The name "Ergodic" refers to the fact that ergodicity of the sequences $\{U_n(a) : a \in A\}$ is what make the discovery of the expected $u$ possible while the Macau Algorithm explores the landscape.[7]

# 4 Simulations

In this section we benchmark the Ergodic Annealing algorithm for two classical combinatorial problems: the Directed Steiner Tree problem on graphs (DST), our initial motivating example, and the Traveling Salesman Problem (TSP), in which the randomness of traffic and viability between two nodes makes compelling the uncertainty in the cost function.

For the DST problem we adapted the Simulated Annealing algorithm of Osborne and Gillett (1991), while for the TSP we adapted the original Simulated Annealing algorithm of Kirkpatrick et al. (1983). As discussed above, by "adapting," we mean replacing the Metropolis routine of the Simulated Annealing algorithm with a Macau routine, leaving the rest unchanged.

## 4.1 Directed Steiner Tree

The first motivating example for the Ergodic Annealing algorithm is the one mentioned in the introduction and pictured in Figure 1, which is formally known as the Directed Steiner Tree problem.

In a DST problem, a directed graph $G(V, E)$ with a non-negative cost $c(e)$ associated to each edge $e \in E$ is considered. The objective is sending a packet from a root node $r$ to each of the terminal nodes $R$, at minimum cost. Each of the $|R|$ packets is allowed to travel through some intermediate nodes, called Steiner nodes. The cost of the whole operation is the sum of the costs of the edges used to send all the packets of information, and the goal is to minimize this quantity.[8] The subset of Steiner nodes used is a variable in this problem, and the optimal configuration coincides with the minimum spanning tree of the subgraph of $G$ induced by the root $r$, the terminals $R$, and an optimal subset of intermediate nodes $I^*$.

Coherently with Figure 1, we consider networks with a *layered structure*, meaning that the vertices can be divided in an ordered partition $\{V_l\}_{l=0}^{L}$, with the singleton $\{r\}$ making up the first layer $V_0$ and the set of terminal nodes $R$ making up the last layer $V_L$. Every edge $e = (v, w)$ in the graph must be such that its vertices are in subsequent layers, i.e. $v \in V_i$ and $w \in V_{i+1}$, for some $i \in \{0, ..., L-1\}$.[9]

---

[7]A stochastic process is ergodic if the time average of a single realization is approximately equal to the ensemble average. Formally,
$$\mathbb{E}[f(U_j)] = \lim_{n \to \infty} \frac{1}{n} \sum_{t=1}^{n} f(U_n) \quad \text{a.s.}$$
for all $j \in N$ and all bounded and Borel measurable $f : \mathbb{R} \to \mathbb{R}$.

[8]Note that we can use an edge as an intermediate channel to reach two different terminal nodes, but its cost will be counted only once. This can be interpreted as having only a fixed "opening" cost of the channel and no capacity constraints.

This is a highly non-trivial combinatorial optimization problem (indeed, it is NP-hard), and Simulated Annealing is a very successful approximation scheme used to tackle it. Therefore a DST with unknown costs is a natural candidate to test the performance of Ergodic Annealing.

A key step required to run an annealing algorithm for a DST problem is the selection of feasible moves from one candidate solution to another.[10] There are different ways to do this, but we decided to follow the proposal of Osborne and Gillett (1991), which simply consists in allowing to move one potential Steiner node ($v \in \bigcup_{l=1}^{L-1} V_l$) from the set of used Steiner nodes to the set of unused nodes, and *viceversa*. Then one can easily compute the new Steiner tree by computing the minimum spanning tree on the resulting subgraph by using Edmonds' algorithm.[11]

To study and compare the performance of Ergodic Annealing with respect to Simulated Annealing we ran both algorithms on a test set of 1000 random graphs with the same true costs —**known** for the Simulated Annealing agent, **unknown** to the Ergodic Annealing one. Each graph $G = (V, E)$ in the test set has 13 layers (so 11 layers of potential Steiner nodes), with a maximum of 12 nodes for each non-root layer. The actual number of nodes for each layer is chosen uniformly at random from $\{2, ..., 12\}$. For each $v \in V$, a node from the previous layer is selected randomly and automatically connected, to guarantee feasibility. All other possible edges in the graph are present with probability 1/2. The (true) arc costs are drawn uniformly from the interval $(0, 1)$, and they are deterministically initialized to $1/2$ for the Macau algorithm.

Since root and terminals are fixed in the optimal Steiner tree, we define the *size* of the graph as the number of potential Steiner nodes. The average size of a graph in the test was 71.3.

In these moderately large graphs, the two algorithms performed quite similarly. Indeed, they reached the same final configuration on 322 graphs, and the average absolute deviation with respect to the best configuration found[12] was 0.04664. In words, on average the two procedures found solutions with costs that differed by 4.66%, sometimes with Simulated Annealing being closer to the true optimal solution, sometimes with Ergodic Annealing performing better.

In the figure below, we present 2 examples of Steiner trees found by Ergodic Annealing and Simulated Annealing on graphs from the test set.

---

[9] Since the graph is directed, the order of the vertices is important.

[10] A move is considered feasible if it transforms a feasible solution into another feasible solution.

[11] In this layered version computing the minimum arborescence is particularly efficient, because the graph is a DAG and there are no recursive calls in the algorithm.

[12] The best configuration is the one of lower cost among the two final configurations found by the algorithms.

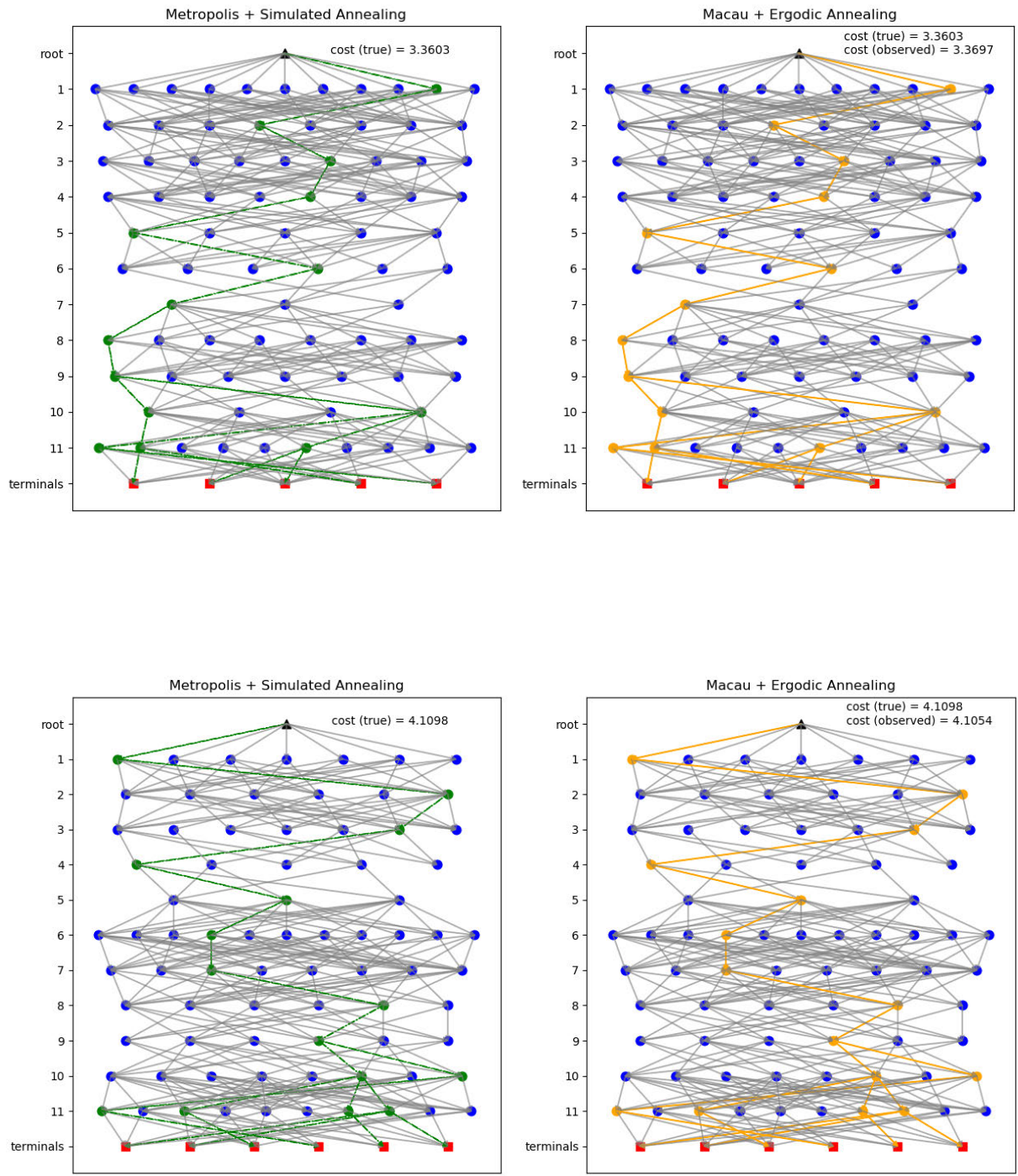

Figure 2: Graphical representation of networks where information travels from a root to the terminals over channels with known or unknown cost.

The simulation, although elementary, supports the conjectures discussed in the previous sections. Indeed, the Macau Algorithm and Ergodic Annealing find configurations of similar cost compared to the ones found by their "cousins" Metropolis Algorithm and Simulated Annealing. This is especially remarkable because Simulated Annealing optimizes over a large but finite set of **known** configuration costs, while Ergodic Annealing searches for a minimum cost configuration on a space that is potentially infinite, because the true costs are **unknown** and are *learned* on a continuous space.

## 4.2 Traveling Salesman Problem

The second benchmark for the validity of Ergodic Annealing studied in this paper is the well-known Traveling Salesman Problem (TSP).

In the classical case a list of cities and distances between each pair of cities are given and known, and the objective is to find the shortest possible route that visits each city and returns to the starting point. Just like the DST, TSP is an NP-hard problem in combinatorial optimization, important in theoretical computer science, operations research and economics.

In our variant, distances are replaced by average travel times, and the fastest route is the objective. The Simulated Annealing algorithm can solve this problem when these travel times are **known**, the Ergodic Annealing can solve it even when travel times are **unknown**.

This time, we ran a simulation over 2000 random instances of TSP, with cities location chosen randomly from the unit square. The number of cities was selected randomly between 30 and 90, with an average size of graphs in the simulation of 59.68. The performance of the two algorithms on the test was almost identical, with Ergodic Annealing performing at least as well as Simulated Annealing on 995 graphs. The average absolute deviation with respect to the best configuration found was 1.90%.

This simulation provides an even stronger evidence than the one found with the previous benchmark about the validity of Ergodic Annealing.

In the figures below, we present 2 examples of optimal routes found by Ergodic Annealing and Simulated Annealing on graphs from the test set.

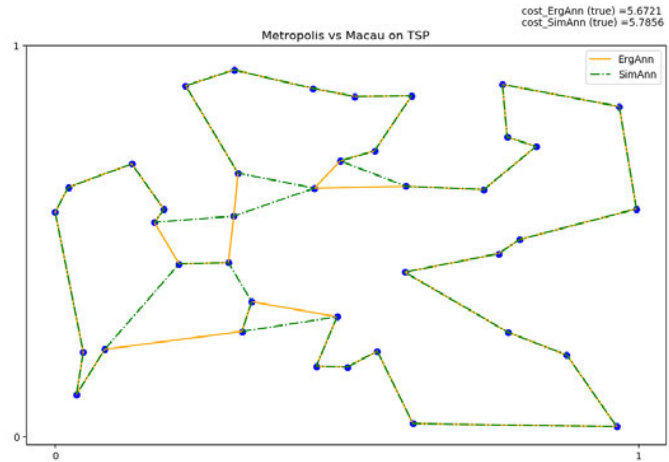

Figure 3: Random TSP instance with 40 cities where Simulated Annealing finds a slightly suboptimal route compared with Ergodic Annealing.

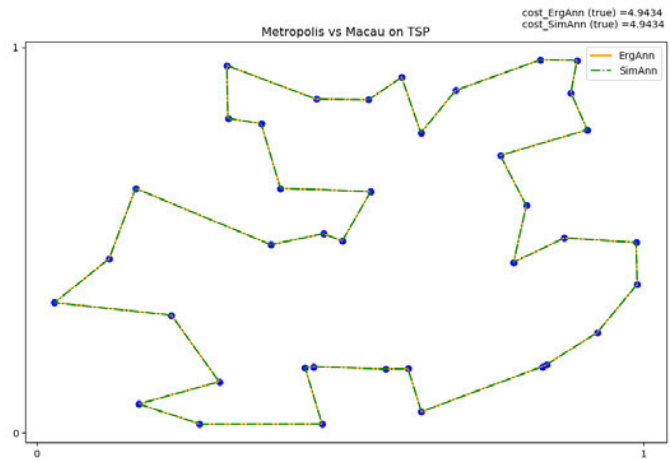

Figure 4: Random TSP instance with 40 cities that produced the same final configuration with both algorithms

# 5 Conclusion

For a given utility function $u$, Ergodic Annealing (implementable by a DM who ignores $u$ and must learn it from the environment) performs almost as well as Simulated Annealing (which requires the DM to know $u$ ex ante). Thus, Ergodic Annealing seems to be a promising extension of Simulated Annealing to decision making under uncertainty.

**Remark 1** *It is important to note that Ergodic Annealing learns the true payoff and optimizes it simultaneously. This speeds up the search because the agent is not interested in finding the true payoff of all alternatives, but only an alternative with highest true payoff. In the simulations this corresponds to the fact that while $u_n$ does not converges to $u$, the payoff of the chosen alternative converges to the optimal payoff of $u$. Our agent is an empirical optimizer, not an empirical statistician.*

# References

[1] Edmonds, J. (1967). Optimum branchings. *Journal of Research of the national Bureau of Standards B*, 71.4, 233-240.

[2] Gelfand, S. B., & Mitter, S. K. (1985, December). Analysis of simulated annealing for optimization. In Decision and Control, 1985 24th IEEE Conference on (Vol. 24, pp. 779-786). IEEE.

[3] Gelfand, S. B., & Mitter, S. K. (1987). Simulated annealing. In Mason, F., Andreatta, G., & Serafini, P. (Eds.). Advanced School on Stochastics in Combinatorial Optimization, CISM, Udine, Italy, September 22-25 1986. World Scientific.

[4] Hastings, W. K. (1970). Monte Carlo sampling methods using Markov chains and their applications. *Biometrika*, 57, 97-109.

[5] Kirkpatrick, S., Gelatt, C. D., & Vecchi, M. P. (1983). Optimization by simulated annealing. *Science*, 220, 671-680.

[6] Madras, N. N. (2002). *Lectures on Monte Carlo methods*. American Mathematical Society.

[7] Metropolis, N., Rosenbluth, A. W., Rosenbluth, M. N., Teller, A. H., & Teller, E. (1953). Equation of state calculations by fast computing machines. *Journal of Chemical Physics*, 21, 1087-1092.

[8] Osborne, L. J., & Gillett, B. E. (1991). A comparison of two simulated annealing algorithms applied to the directed Steiner problem on networks. *ORSA Journal on Computing*, 3, 213-225.

[9] Romeo, F., & Sangiovanni-Vincentelli, A. (1991). A theoretical framework for simulated annealing. *Algorithmica*, 6, 302-345.